\documentclass[runningheads]{llncs}
\usepackage{graphicx}
\usepackage{amsmath,amssymb} % define this before the line numbering.
\usepackage{booktabs}
\usepackage{color}

\usepackage{multirow}

\usepackage{booktabs}
\usepackage{threeparttable} 

\usepackage{tabu}

\usepackage[width=122mm,left=12mm,paperwidth=146mm,height=193mm,top=12mm,paperheight=217mm]{geometry}

\mainmatter
\def\ECCV18SubNumber{1754}  % Insert your submission number here

\title{Spatial-Temporal Synergic Residual Learning for Video Person Re-Identification} 

\titlerunning{}

\authorrunning{}

\author{
  Xinxing Su$^{1*}$ \hspace{0.7em} Yingtian Zou$^{1*}$ \hspace{0.7em} Yu Cheng$^2$ \hspace{0.7em} Shuangjie Xu$^1$ \\ Mo Yu$^3$ \hspace{0.7em} Pan Zhou$^1$ 
}
\institute{$^1$ Huazhong University of Science and Technology,\hspace{1em}   $^2$Microsoft AI \& Research\\
$^3$IBM Research AI\\
  ($^*$ denotes equal contribution)}

\begin{document}

\maketitle

\begin{abstract}
% We tackle the problem of person re-identification in video setting, which has been viewed as a crucial task in many applications. Meanwhile, it is very challenging since the task requires learning an effective representation from video sequences with heterogeneous spatial-temporal information. In this paper, we present a novel Spatial-Temporal Synergic Residual Network (STSRN) to simultaneously extract the generic and specific spatial-temporal features of consecutive frames. 
% % Specifically, we propose a novel spatial residual extractor temporal residual processor and the
% % In additionally, the proposed model can leverage temporal information to keep smooth spatial alignments between consecutive frames. STRL is very parameter efficient and provides an end-to-end framework for training and prediction. 
% Extensive experiments are conducted on several large scale datasets MARS, PRID2011 and ILIDS-VID. The results demonstrate that the proposed method achieves consistently superior performance and outperforms all of the state-of-the-art methods. \footnote{Our code will be released soon.}
We tackle the problem of person re-identification in video setting in this paper, which has been viewed as a crucial task in many applications. Meanwhile, it is very challenging since the task requires learning effective representations from video sequences with heterogeneous spatial-temporal information. We present a novel method - Spatial-Temporal Synergic Residual Network (STSRN) for this problem. STSRN contains a spatial residual extractor, a temporal residual processor and a spatial-temporal smooth module. The smoother can alleviate sample noises along the spatial-temporal dimensions thus enable STSRN extracts more robust spatial-temporal features of consecutive frames. Extensive experiments are conducted on several challenging datasets including iLIDS-VID, PRID2011 and MARS. The results demonstrate that the proposed method achieves consistently superior performance over most of state-of-the-art methods.
\end{abstract}

\section{Introduction}
Person Re-identification (ReID) is the problem of associating different tracklets of a person across non-overlapping cameras. It has become increasingly popular for its crucial applications in visual surveillance and human computer interaction. Benefited from tremendous success of deep learning, the computer vision field has witnessed the prominent progresses in image-based person re-ID~\cite{liao2015person,zhao2013unsupervised,ahmed2015improved}, which only utilizes the spatial information. However, single-shot appearance features of people are intrinsically limited to the inherent visual ambiguity. More recently, many attentions have been shifted to the video re-ID since its natural setting and some benefits with sequential information. 

Video re-ID faces several significant challenges, like cluttered backgrounds, out-of-focus targets, misalignments and large appearance changes as a person moves between cameras. In a meanwhile, how to extract more comprehensive representations, particularly incorporate spatial and temporal information available in videos, is still under-studied. To overcome the these issues, recent video-based methods have tended to utilize RNNs~\cite{rumelhart1986learning} (or CNN-RNNs) to take consecutive frames as inputs, and adaptively incorporate temporal information~\cite{mclaughlin2016recurrent,yu2017three,chung2017two,xu2017jointly,zhu2016video,dai2018video,zhou2017see}. For instance,~\cite{zhou2017see,xu2017jointly} focus on considering the mutual influence between video sequences. On the other hand, another frequently used strategy is the multi-shot matching~\cite{bazzani2010multiple,zheng2016mars,farenzena2010person,liu2017quality}, where they only utilize convolution-based representations. Pooling operation in these methods aggregates frame-level features into a global vector, which has demonstrated its simplicity but effectiveness. However, the residual learning in CNN or RNN is rarely studied in the literature related to person ReID. 

To tackle with aforementioned issues, we propose Spatial-Temporal Synergic Residual Network (STSRN), a novel method aiming at solving the coherent representation learning bottlenecks on existing spatial-temporal models. This is achieved by exploring three different modules: a spatial residual extractor, a temporal residual processor and a spatial-temporal smoother. Particularly, the spatial residual extractor can extract discriminative frame-level spatial features while temporal processor could further improve the ability of modelling long-range dependencies and eliminate redundancy in videos with the help of residual learning. The spatial-temporal smoother makes a gentle transition between spatial domain and temporal domain. 

\begin{figure}[t]
\centering
\includegraphics[width=12cm]{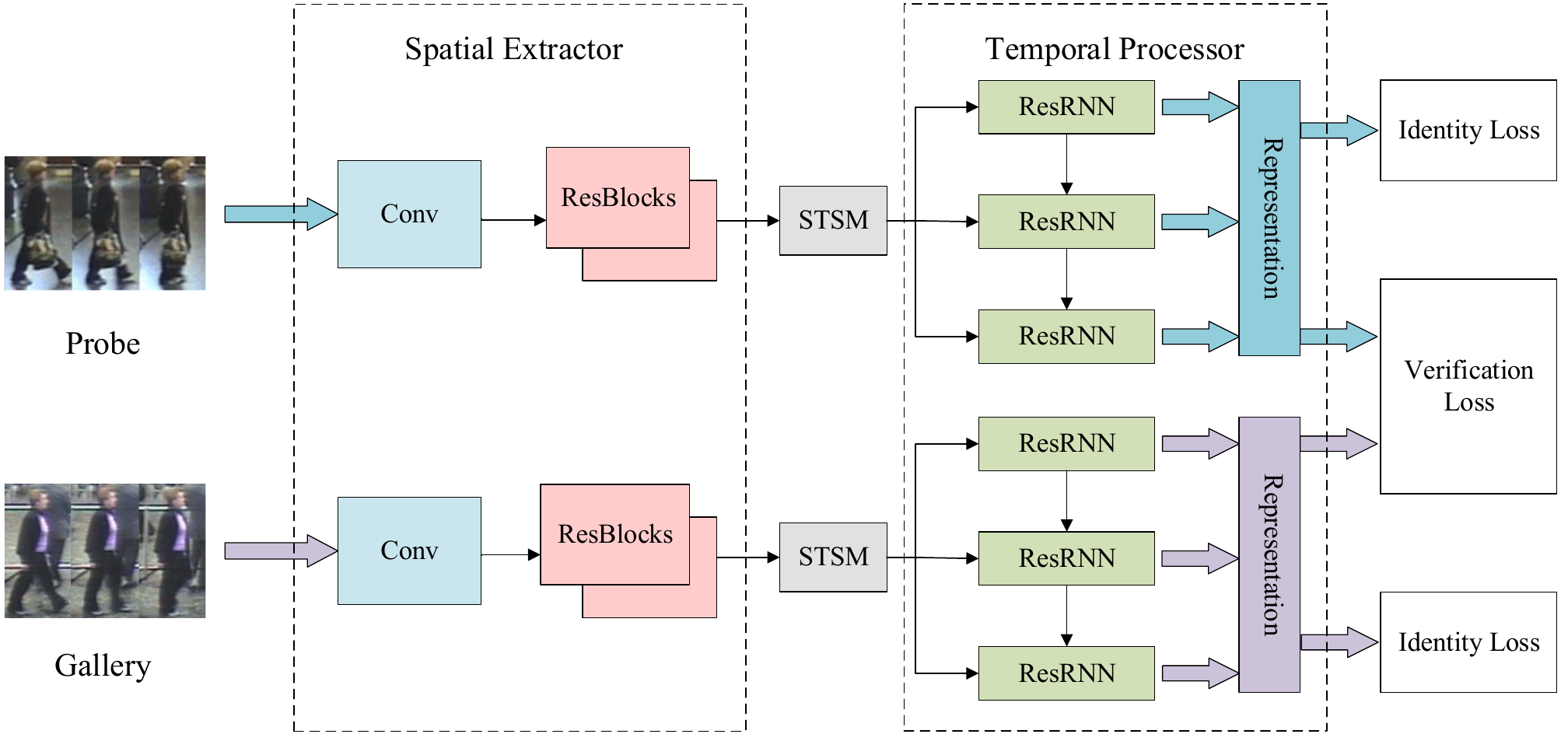}
\caption{Overview of Spatial-Temporal Synergic Residual learning Network (STSRN).  The framework extracts the generic and specific spatial-temporal features of consecutive frames with the help of synergic residual learning between spatial residual blocks, spatial-temporal smooth module (STSM) and residual RNN (ResRNN)}
\label{fig_simple}
\end{figure}

Moreover, our proposed model provides an end-to-end and parameter-efficient solution. Instead of extracting frame-level representations by using GoogLeNet~\cite{szegedy2015going} such as in~\cite{liu2017quality} and~\cite{dai2018video}, or time-consuming in inference stage~\cite{zhou2017see,xu2017jointly}, we employ a much smaller architecture in that STSRN could be more likely to apply to real-time video surveillance system.

In summary, our main contributions are in three-folds:
\begin{itemize}
\item We propose the Spatial-Temporal Smooth Module (STSM) for straining to eliminate the noise interference in embeddings when transformed from spatial domain into temporal domain. STSM demonstrates its effectiveness theoretically and experimentally in spatial-temporal synergic residual learning.
\item We devise a new spatial-temporal network which has a better representation than other CNN-RNNs. Compared with most previous approaches, STSRN is very parameter saving and efficient for both training and prediction.
\item We evaluate STSRN on three popular video benchmark datasets, iLIDS-VID, PRID2011 and MARS. The experimental results show that our model beat most of state-of-art approaches with much fewer parameters. To prove the validity of model, we also perform cross-dataset testing on PRID2011. 
\end{itemize} 

\section{Related Work}
%Vanilla method is adopt general deeper models such as VGG, ResNet34, ResNet101. But
%\subsection{Video-Based Person Re-ID}
Person re-identification has been a hot topic. Especially in recent years, video-based person re-ID task has drawn increasing attention due to its practicability in surveillance. In video-based re-ID, each tracklet contains a number of frames rather a single image. Comparing with single shot person re-ID, video-based fashion utilizes more potential information thus has a better prospect.
Previous work on video person re-ID can be roughly categorized into
two classes: hand-crafted feature based methods \cite{hadsell2006dimensionality,hamdoun2008person,wang2014person,liu2015spatio,zhu2016video,you2016top,zhang2017video} and deep learning based methods \cite{ahmed2015improved,xu2017jointly,zheng2016person,he2016deep,mclaughlin2016recurrent,chung2017two,liu2017quality,dai2018video,zheng2016mars}. 

Classical separate solutions target at feature extraction \cite{hadsell2006dimensionality,hamdoun2008person,wang2014person} and distance metric learning~\cite{zhu2016video,you2016top}. For feature extraction, Wang $et~al.$~\cite{wang2014person} represent video fragments by the HOG3D features and the average color histograms, where they screen these discriminative fragments from noisy sequences by estimating the flow energy profile (FEP). By distance metric learning, Liu ${et~al.}$~\cite{liu2015spatio} consider spatial alignment and temporal alignment to treat the appearance of different body parts independently. They then represent feature using fisher vector. 

Deep methods can be further subdivided into two aspects: (1) The methods focusing on spatial image-level representations~\cite{liu2017quality,zheng2016mars}, which generally build tremendous and complex networks. For example, Liu $et~al.$~\cite{liu2017quality} directly learn to automatically score the image data according to its quality.
% Dai $et~al.$~\cite{dai2018video} make use of GoogleNet~\cite{szegedy2015going} to learn the spatial representation and BiLSTM~\cite{schuster1997bidirectional} to learn the temporal residual. Chung $et~al.$~\cite{chung2017two} apply two stream Siamese Network~\cite{hadsell2006dimensionality} to extract image features and motion features  simultaneously.
Zheng \emph{et al.}~\cite{zheng2016mars} tries to train a classification network~\cite{jia2014caffe} where each single image is represented by a feature vector.
%However, these methods thoroughly learn image-level semantic information by convolutional-based extractor. 
It is inevitable to establish a deeper and complicated hierarchy in their models. 
%Our succinct methodology build on a simple network which is  easier to deploy. In the meanwhile, our model achieves the state-of-the-art.
(2) The ones paying more attention to temporal sequence-level representations, most of which adopt CNN-RNNs~\cite{mclaughlin2016recurrent,zheng2016mars,dai2018video,chung2017two,yu2017three,xu2017jointly,zhou2017see}. Pooling based strategies in these models aggregate the features of tracklet which have better scalability in the sequence level. Dai $et~al.$~\cite{dai2018video} employ double GoogLeNets~\cite{szegedy2015going} to extract general features and aligned features, which are processed by double Bi-LSTMs~\cite{schuster1997bidirectional} separately. It also propose a spatial-temporal transformer network (ST$^2$N) to fuse different level features from the a GoogLeNet.
%As the word suggest, drawing support from temporal attention strategy is a popular way. 
Xu $et~al.$~\cite{xu2017jointly} propose an jointly spatial-temporal attention matrix to maximum the relativity of different scene feature. 
% Zhou $et~al.$~\cite{zhou2017see} 
% % applies the jointly spatial-temporal attention learning. He 
% propose the temporal attention model (TAM) to focus on discriminative frames, which is jointly learned with the spatial recurrent model (SRM) to integrate the surrounding information at different spatial locations for better similarity evaluation. 
Chung \emph{et al.}~\cite{chung2017two} process spatial information and motion information separately by SpatialNet and TemporalNet.
%Considering for the practicability, we also aim at obviating the computational burden for faster retrieving in this paper. 

As predicted in the Zheng \emph{et al'}s survey~\cite{zheng2016person},
%discriminative combination of appearance and spatial-temporal models is an effective solution in future video re-ID research.
recent multi-shot based re-ID methods reporting competitive accuracy mainly adopt the discriminative combination of appearance and spatial-temporal models or directly fine-tuned an identification model pretrained in ImageNet~\cite{deng2009imagenet}. With the increase in capacity of spatial representation, they intensify the risk of over-fitting and time complexity. Meanwhile, deep residual learning has achieved tremendous success in various visual tasks~\cite{he2016deep,yxlcvpr17,ballas2015delving,feichtenhofer2016spatiotemporal}. In Feichtenhofer \emph{et al.}' work~\cite{feichtenhofer2016spatiotemporal}, it builds on injecting residual connections from appearance into motion pathway in a two-stream convolutional networks. Utilizing additive merging of signals can ease the training of network which has an advantage of informative tasks like video analysis. Instead of fusing different level features for learining aligned features~\cite{dai2018video}, we devise a unidirectional spatial residual extractor to perform progressively fusion layer by layer.
For the purpose of a better representation with parameter saving fashion than other CNN-RNNs, we propose a spatial-temporal synergic residual network which can reduce the both information redundancy and noisy interference. 
%To overcome the gradient vanishing, He \emph{et al.}~\cite{he2016deep} proposes shortcuts in layer connection.  Utilizing additive merging of signals can ease the training of deeper network which has an advantage of informative tasks like video analysis. Residual learning has got extensive applications in video analysis~\cite{zhang2017deep,ballas2015delving,feichtenhofer2016spatiotemporal}.  
%\subsection{Deep Residual Learning}
%%%%%%%%%%%%%%%%%%%%%%%%%%%%%%%%%%%%%%%%%%%%%%%%%%%%%%%%%%%%%%%%%%%%%%%%%%
%%%%%%%%%%%%%%%                     3            %%%%%%%%%%%%%%%%%%%%%%%%%  %%%%%%%%%%%%%%%%%%%%%%%%%%%%%%%%%%%%%%%%%%%%%%%%%%%%%%%%%%%%%%%%%%%%%%%%%%

\begin{figure}[t]
\centering
\includegraphics[width=12cm]{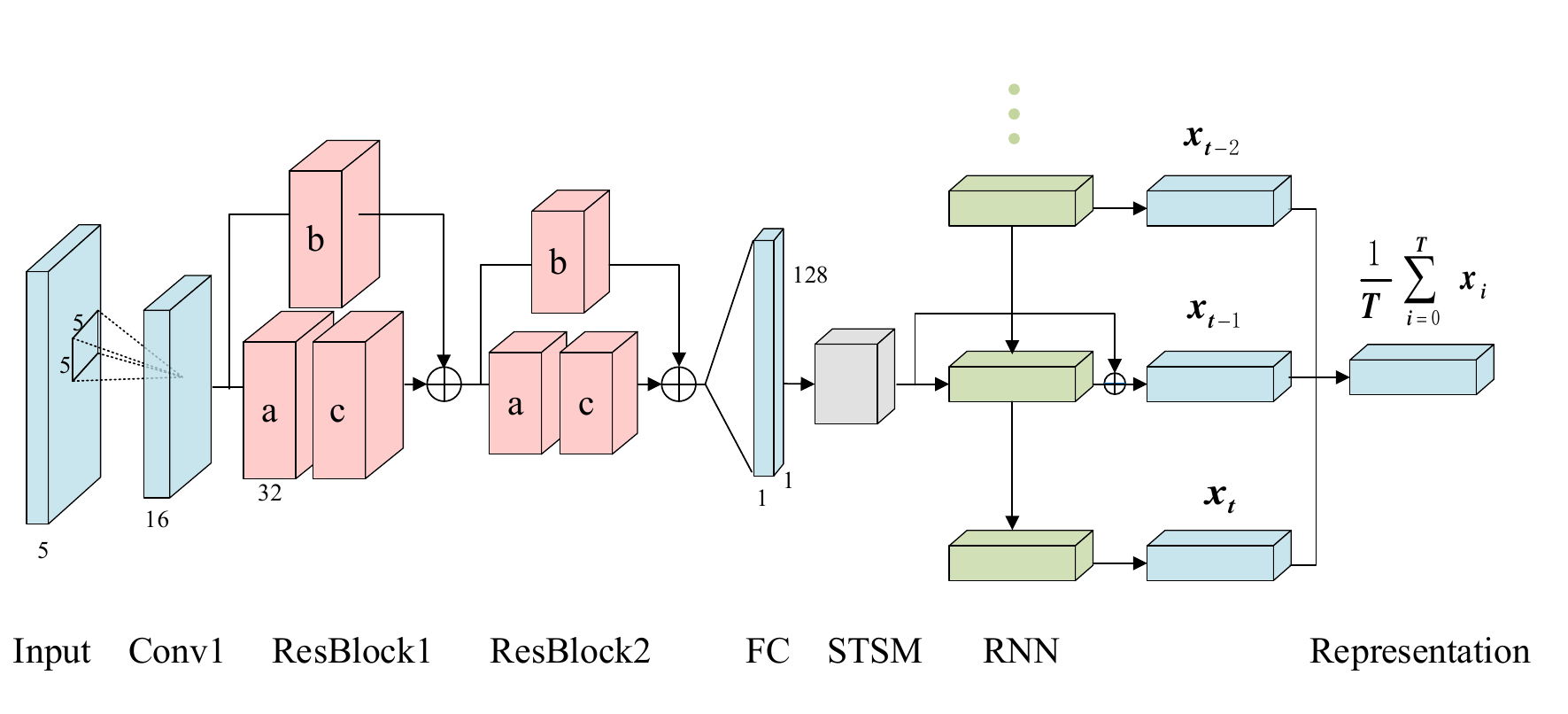}
\caption{Details of the network structure. Kernel size, stride and padding of cube ``a, b" are 5, 4 and 1, respectively. Kernel size, stride and padding of cube ``c" are 3, 1 and 1, respectively. ``$\oplus$'' is element-wise addition.}
\label{fig_3dmodel}
\end{figure}

\section{The Proposed Model Architecture}
\subsection{Overview}
Our proposed framework, STSRN, belongs to the spatial-temporal models. As shown in Figure~\ref{fig_simple}, our STSRN is trained as a Siamese network architecture~\cite{hadsell2006dimensionality} by passing a pair of image sequences, each of which is a slice of the tracklet. After the fully-connected layer, it produces an embedding for every image and then be smoothed by the STSM. For these embeddings, the temporal residual processor would generate the frame-level features, which will then be aggregated into the sequence-level representation. Overall, the network outputs two global vectors for computing the Euclidean distance between them and predicting the probability distribution over training identities.

There are two crucial parts and an interlayer between them. The first one lies in residual blocks, which could learn more discriminative features than the plain counterpart. Compared with the standard forms of the ResNet~\cite{he2016deep}, our residual blocks are more suitable for re-ID tasks on hand, as will be explained below. Introducing residual learning into RNN brings further performance gains and robustness. Meanwhile, the smoother guides the temporal processor to fucus on the most relevant frames for matching with a novel gating mechanism.
% further make use of complementary cues about the person identity in the other hand.
% Instead of stacking several convolutional layers per residual block and further grouping these blocks together as an layer of ResNet, pruned 
% spatial feature extractor combines single convolutional layer and nstead of using three convolutional layers and a vanilla RNN~\cite{mclaughlin2016recurrent}, for extracting spatial features and exploiting shortcut connections in the sequential information aggregation network for temporal residual learning. To address the shortcomings we pointed above, we further propose a Spatial-Temporal Alignment Module (STAM) inserted between spatial feature extractor and temporal feature processor to reduce the misalignments. 

\begin{figure}[t]
\centering
\includegraphics[width=11cm]{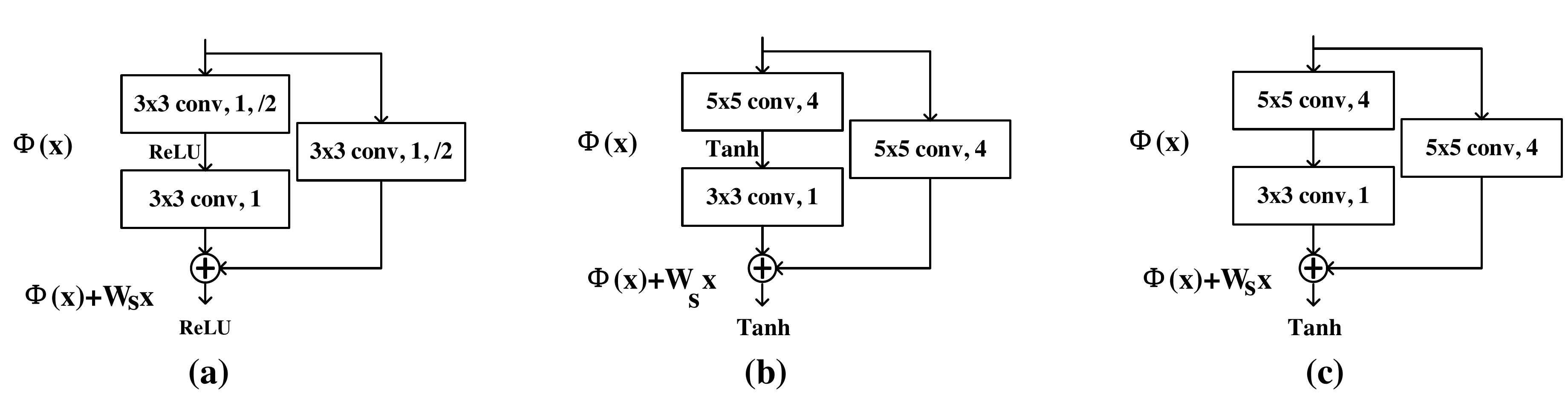}
\caption{Comparisons on different structures of residual blocks. The digits in rectangle indicate kernel size, layer type, padding and stride (if exists), respectively. (a) Residual connection in ResNet~\cite{he2016deep}, (b) Variant of $(a)$ with larger kernel size and padding, and Tanh activation, (c) The same architecture as $(b)$ without in-block activation}
\label{fig_resblock}
\end{figure}

\subsection{Spatial Residual Extractor}
\label{ssec:spatial_extractor}
Recently, the state-of-the-art multi-shot re-ID methods tend to adopt complex hierarchies such as Two Stream Siamese Networks~\cite{chung2017two}, QAN~\cite{liu2017quality} and TRL~\cite{dai2018video}.
%the discriminative combination of appearance and spatial-temporal models or directly fine-tuned a classification model pretrained in ImageNet~\cite{deng2009imagenet}. make full use of their advantages comprehensively
With more learnable parameters, these methods have potential superiority in terms of model capacity but leads to higher risk of over-fitting and time complexity.
This section proposes a new spatial feature extractor, which both simplifies the model topology and improves the discriminability of the representation for video-based re-ID task.
% seeking a more comprehensive characterization. Thus we devised a new spatial feature extractor displayed in Fig.\ref{fig_3dmodel}

From Figure~\ref{fig_3dmodel}, we can see that the spatial extractor can be divided into three submodules: one for convolutional and the others for residual submodules. Specifically, each submodule is composed of filter banks, hyperbolic-tangent (Tanh) activation-function and max-pooling. 
The input to the network consists of both optical flow channels and colour channels. Given the input sequences $S = \{I^{\tau}\in \mathbb{R}^{w\times h\times c} | \tau = 1,...,T\}$, the first submodule outputs feature maps $C^{'}(S)$, where $T$ is the sampling length of sequence and   
%In like manner, we can get gallery input $ S'=\{I'^{(1)},I'^{(2)},...,I'^{(T)}\}$ sampled from another camera. 
$C^{'}(S) = Tanh(Maxpool(Conv(S)))$.

As shown in Figure~\ref{fig_resblock}, instead of applying the standard form from ResNet~\cite{he2016deep}, we adopt $c$ as our residual module for three purposes. Firstly, we use the Tanh rather than Rectified Linear Unit (ReLU) in residual blocks to keep the data distribution compatible with subsequent recurrent layer. Secondly, removing the activation layer inside the residual block alleviates the gradient vanishing problem caused by Tanh. Last but not the least, this layout has larger kernel size and padding, which are more applicable on low-resolution video ReID task. Compared to~\cite{mclaughlin2016recurrent}, the introduction of residual blocks offers more paths for information flowing from early layer to later layer.
% $f_{conv}(S) = Tanh(Maxpool(Conv(S)))$.
% Analogously, the feature sets of the first residual submodule can be written as $f_{res}^{1}(f_{conv}(S))$. 
%So we apply this conception to our task-specific network which is more suitable for low-resolution input, see Figure~\ref{fig_resblock}.for one of our intentions is to overcome computational burdens caused by deep neural networks
Formally, the feature set of the residual submodule can be formulated as follow:
\begin{align}
   f_{res}^{k}(x) &= \Phi(x) + W_s x   \label{eq_spatialResBlock1}
\end{align}
% $$f_{res}^{k}(x) = \Phi(x) +  W_s x$$
% $$f_{res}^{1}(C^{'}(S)) = \Phi(C^{'}(S)) +  W_s C^{'}(S)$$
where $x$ and $k$ represents the input to to the block, and the index of blocks, respectively. Next, the embeddings 
$f_s(S) = \{f_{s}(I^{\tau})\in \mathbb{R}^{N}, \tau = 1, ..., T\}$
% [f_{s}(I^{(1)}),f_{s}(I^{(2)}), ... ,f_{s}(I^{(T)})] \in \mathbb{R}^{N\times T}$ 
are mapped by the fully-connected layer, where $N$ is the dimensionality of embeddings and $f_s$ a series of non-linear mapping from the raw pixels to a new embedding space.
% Besides, the Section 4.4 will illustrate how did we explore our final model in detail from various combinations of residual blocks.
% $I^{(t)}$ is the image of slice at time t. $L$ is the number of residual blocks in spatial feature extractor. 
% Before this model, we have tried many solutions such as margin penalty, fully convolutional networks. They bring the problems like unstable, over-fitting and slow convergence etc. Therefore, the Section 4.4 will illustrate how we explore reasonable model structure in detail. 
% To simplify description, we denote the convolutional submodule at $l$-th of the net as $f_{conv}^l(x)$.  Similarly, $f_{res}^l(x)$ indicates the residual operation at $l$-th submodules.  Note that, there are some differences from ResNet~\cite{he2016deep}, the ``residual blocks" in our model are not the standard form. But we apply this conception to our task-specific network which is more suitable for low-resolution input, see Figure~\ref{fig_resblock}. 
%We expect our model can learn the residual information without complicating the topologies.???? Sticking to this principle, we build our network by single stream with ,shown in Fig.\ref{fig_resblock}.
\subsection{Spatial-Temporal Smooth Module}
Because of the challenges such as occlusions and background clutter in video person re-ID, it is necessary to simultaneously make full use of space and temporal information~\cite{zheng2016person}. 
% However, none of the previous research could deal with the problem very well. 
For this, a Spatial-Temporal Smooth Module (STSM) is introduced to discard noisy motion occurring over a short period through selecting most salient spatial features temporally.

% Especially, when large gait, illumination changes between adjacent frames the when one of the jitter, blur, occlusions in a frame. 
Following Section \ref{ssec:spatial_extractor}, we obtain spatial embedding set $f_s(S)$, whose adjoining embeddings are then fused by learnable parameters $\theta$ and $\omega$. More specifically, the sum of each element of $\theta$ and $\omega$ is subject to one. Therefore, the smoother aggretates $f_s(I^{\tau-1})$ and $f_s(I^{\tau})$ temporally as follow:
\begin{align}
   f_{s\to t}^{\tau} &= \theta \odot f_s(I^{\tau}) + \omega \odot f_s(I^{\tau-1}) \label{eq_STSM1} \\
   & = f_s^{\tau} + \omega \odot (f_s^{\tau-1} - f_s^{\tau})
\end{align}
where $\omega \in \mathbb{R}^{N}$ and $f_{s\to t}^{1} = f_s^{1}$. 
% \begin{align}
% f_{s\to t}(I^{(1)}) = f_s(I^{(1)}) \label{eq_STSM2}%\\
% \end{align}
% $$f_{s\to t}(I^{(t)}) =  U' f_s(I^{(t)}) + V' f_s(I^{(t-1)}),~~f_{s\to t}(I^{(1)}) =  f_s(I^{(1)})$$
$\omega$ serve as gate mechanism: when some frame is interference, the mechanism would balance the ratio of information flow, thus guiding the recurrent layer to fucus on more discriminate features.
%That is, the misalignments between spatial representation and temporal representation which is why we propose the spatial-temporal alignment module. Following Section 3.2 and 3.3, we obtain spatial and temporal representation $f_s(S)$ and $f_a(S)$, respectively. Generally, we usually set the first hidden state $r^{(0)}$ in Eq.~\ref{eq_rnn1} of RNN to $\boldsymbol{0}$ where $\boldsymbol{0}$ is the zero vector. Then we can only get the wighted spatial representation $o^{(0)} = U x^{(0)}$. If we take use of the $o^{(0)}$ in (\ref{eq_rnn2}), it will cause the fitting error. Aiming at this issue, we devise the Spatial-Temporal Alignment Module (STAM) to eliminate the misalignment between the spatial representation and temporal representation. In this paper, we only need to fix the linear fitting error. Our specific solution is to add a bias $b_{s\to t}$ like this:

\subsection{Temporal Residual Processor}
Inspired by the residual spatial connections, we also consider to employ residual learning temporally to further help the re-ID task. As shown in Figure~\ref{fig_3dmodel}, we introduce a shortcut connection from the output of STSM to the output of the RNN. On one hand, RNN allows informative data to propagate from the first time step to the last one and accumulates discriminative information along the temporal dimension. On the other hands, through alternative information of current frame, the residual connection can prevent redundant information from
worsening the temporal accumulation, which makes it more robust to noisy motion. Furthermore, this novel fashion is parameter-free.

For simplicity, we reformulate the embeddings produced by fully-connected layer as $[x^{1},x^{2}, ... ,x^{T}]$, and then obtain smoothed embedding matrix $M$ transited by STSM:
\begin{align}
   M &= [\hat{x}^{1},\hat{x}^{2}, ... ,\hat{x}^{T}],~~\hat{x}^{\tau} = f_{s\to t}^{\tau} \label{eq_outSTSM1}
\end{align}
Thus, we can incorporate recurrent connections
between the STSM and temporal-pooling layer as
follows:
\begin{align}
   \hat{o}^{\tau} &= U\hat{x}^{\tau} + Vr^{\tau-1}  \label{eq_rnn1} \\
   o^{\tau}&= \frac{\hat{o}^{\tau} + \hat{x}^{\tau}}{2}, r^{\tau}= Tanh(o^{\tau}) \label{eq_rnn2}
\end{align}
where $r^{\tau}\in \mathbb{R}^{L}$ is the hidden state at time $\tau-1$, and $o^{\tau}\in \mathbb{R}^{L}$ is the output at time $\tau$, $U \in \mathbb{R}^{N\times L}$ and $V \in \mathbb{R}^{L\times L}$ are the projection weights for the observation $\hat{x}^{\tau}$ and previous hidden state, respectively. Next, all outputs of the residual RNN are performed mean-pooling over 
the temporal dimension to produce a global feature vector $v_{g}$ characterizing the whole input:
\begin{align}
	v_{g} = \frac{1}{T}\sum\limits_\tau o^{\tau} \label{eq_rnn_out}
\end{align}

Our residual connections in RNN provide alternative signals, which make the training of RNN an easier problem, i.e., the features are smoothly transferred across video sequence. Besides, it further reduces the information redundancy beyond the benefits of STSM.
% On one hand, residual learning with alternative signals can be regarded as an attention mechanism in space and time cross domain. On the other hand, this connection mode can alleviate the pressure of training RNN where the representation is smoothly transferred from space to time. In a nutshell, we reduce the information redundancy of the RNN. In later section, we experiment with temporal residual learning to demonstrate our temporal processor is more stable in optimizing.
% During backpropagation phase, the criterion will compute the temporal residual error: $error_S = y^{(S)} - f_t(S)$
% That's how we apply the residual learning in RNN. 
%\subsection{STAM for Synergic Residual Learning}
% $V^{(0)}$
\subsection{Training Objective}
Given a pair of input sequences ($S^m$, $S^n$) of persons $m$ and $n$, our STSRN produces two global spatial-temporal vectors ($v_g^m$, $v_g^n$). Note that $S^m$ and $S^n$ could be of different lengths. We define our training objective on the joint identification and verification loss from~\cite{sun2014deep}. The verification cost, i.e. siamese cost, tries to minimize the distance between $v_g^m$ and $v_g^n$ when they belongs to the same identity and maximize the distance otherwise:
\begin{equation}
\mathcal{L}_{veri}(v_g^m, v_g^n)=\left\{
\begin{aligned}
&\Vert v_g^m - v_g^n\Vert_2^2~, &p=g\\
&max( 0 ,~m - \Vert v_g^m - v_g^n\Vert_2 )~, &p \neq g
\end{aligned}
\right. \label{eq:siameseLoss}
\end{equation}
where $m$ is the margin. Besides, we apply the cross-entropy loss to obtain the identity cost on persons $p$ and $g$ as $\mathcal{L}_{iden}(v_p)$ and $\mathcal{L}_{iden}(v_g)$. The total training objective is the sum of these cost.

% At the top of the net, we apply two types of criterion -- verification loss and identity loss to jointly optimize the model. Our formulaic description is:
% \begin{equation}
% \mathcal{L} = \mathcal{L}_{id}(S_p) + \mathcal{L}_{iden}(S'_g) + \mathcal{L}_{ve}(S_p,S'_g) \label{eq:totalLoss}
% \end{equation}
% \begin{equation}
% \mathcal{L}_{id}(S_p) = \sum\limits_{p\in T} log \label{eq:idenLoss}
% \end{equation}
%Simultaneously, the dual path of the siamese network get the same form representation $f_t(S')$ as another of the input pair. $S' = \{I'^{(1)},I'^{(2)},...,I'^{(t)}\}$ has the same length $t$ with the $S$ where $S'$ and $S$ are the pairwise sequence examples.

\section{Experiments}
This section empirically compares our proposed framework with previous state-of-the-art methods for video-based person re-identification on three popular benchmarks: iLIDS-VID~\cite{wang2014person}, PRID2011~\cite{hirzer2011person} and MARS~\cite{zheng2016mars}.
We also conduct analysis to better understand the effects of several crucial components and parameter settings.

\subsection{Datasets}
\subsubsection{iLIDS-VID dataset.}
The iLIDS-VID dataset~\cite{wang2014person} consists of 300 distinct pedestrians with one pair of sequences from two non-overlapping camera perspectives for each person. It was captured at a crowded airport arrival hall with significant background clutter, extremely heavy occlusion and viewpoint/illumination variations across camera views, which makes it one of the most challenging datasets used for multi-shot person re-ID task. The length of each image sequence varies from 23 to 192, with an average number of 73 images.
\vspace{-0.4cm}
\subsubsection{PRID2011 dataset.}
Although also featuring multiple person trajectories from two different, static surveillance cameras similar to the former, PRID2011~\cite{hirzer2011person} has different number of identities for Camera A and Camera B, respectively. Only the first 200 persons appear in both views, which results in 400 image sequences totally, and it was captured in uncrowded outdoor scenes with relatively simple backgrounds and rare occlusions. The length of consecutive frames in single camera view for each person ranges 5 to 675, with an average number of 100.
\vspace{-0.4cm}
\subsubsection{MARS dataset.}
The Motion Analysis and Re-identification Set (MARS)~\cite{zheng2016mars} is a large-scale video dataset which contains 1,261 different identities in over 20,000 tracklets. These tracklets are automatically generated by the pedestrian detector DPM and tracker GMMCP making it more realistic and challenging than datasets above. Most IDs are captured by 2-4 cameras and camera-2 produces the most tracklets. Most tracklets contain 25-50 frames, and there are 13.2 tracklets on average for each identity.

\subsection{Implementation Details}
For these experiments, we randomly split each dataset into two non-overlapping subsets with same amount of identities for training and testing, respectively. The results are reported using the average \emph{Cumulative Matching Characteristics} (CMC) curves under ``10-fold cross validation''. In the testing phase, the probe set and the gallery set contain data from two different cameras for iLIDS-VID and PRID2011; and we only use the first 200 identities appeared in both cameras for PRID2011. As for MARS, we randomly chose 2 cameras of the same person out of the ensemble following~\cite{xu2017jointly}. For the fairness comparison, we set the sampling length of each person sequence to 16 and 128 for training and testing, respectively.

Data preprocessing contained several steps~\cite{xu2017jointly}: The optical flow channels were calculated between each pair of images horizontally and vertically by Lucas-Kanade algorithm~\cite{lucas1981iterative} then normalized to fall within the range -1 to 1 while the RGB images were converted to YUV color space. Both the RGB channels and flow channels were normalized to the range of [0,1], after which an another normalization operation was applied to have zero mean and unit variance channel by channel to keep the consistency of the influence of each feature on the objective function.

The training sequences were augmented in the form of randomly cropping and mirroring, which was applied to all frames of a given sequence to improve the ability of generalization. We also adopted the same augmentation step in test phase. Following~\cite{xu2017jointly}, image sequences from different cameras of same pedestrian were considered as positive pairs while those of different pedestrians negative pairs. The positive and negative pairs were alternatively fed into the network.

The initialization of weight parameters for convolutional layers and fully-connected layers was done by Xavier method~\cite{glorot2010understanding}. The hyper-parameters were set based on~\cite{mclaughlin2016recurrent} except that init learning rate was set to 2 ${\times}$ $10^{-3}$. With the stochastic gradient descent (SGD) algorithm, the proposed model was trained on NVIDIA GTX-1080Ti GPUs.
% We followed~\cite{wang2014person} to adopt a common evaluation mechanism -- Cumulative Matching Characteristics (CMC) curve which indicates the probability of finding the correct match in the top N matches within the ranked gallery. Under this evaluation, we calculated the top-1,5,10,20 curve respectively, shown in Table~\ref{tab:performance_cmc}. All tests are repeated 8 times and the average accuracy rates is reported to ensure statistically reliable evaluation.
\begin{table}[t]
  %\centering  
  %\fontsize{6.5}{8}\selectfont  
  \begin{threeparttable}
  \caption{Comparisons of our network with other state-of-the-art methods on iLIDS-VID and PRID2011 in terms of CMC rank-1, rank-5, rank-10 and rank-20 (\%).}  
  \label{tab:performance_cmc}
  	\begin{tabular*}{\textwidth}{l@{\extracolsep{\fill}}*{9}{c}}
    %\begin{tabular}{cccccccccccccc}
    \toprule  
    %\multirow{2}{*}{Dataset}&
    \multicolumn{2}{c}{\multirow{2}{*}{Model}}
    &\multicolumn{4}{c}{iLIDS-VID}&\multicolumn{4}{c}{PRID2011}\cr
    %\multicolumn{3}{c}{ 1-shot}&\multicolumn{3}{c}{5-shot}\cr  
    %\cmidrule(lr){2-4} \cmidrule(lr){5-7}  
   	\cmidrule(lr){3-6}\cmidrule(lr){7-10}
    %&CelebA&AwA&CUB
    &&R1&R5&R10&R20&R1&R5&R10&R20\cr  
    \midrule 
%     \multicolumn{2}{c}{TopRank}  &22.5&56.1&72.7&85.9&31.7&62.2&75.3&89.4\cr
    \multicolumn{2}{c}{VR\cite{wang2014person}}
    &34.5&56.7&67.5&77.5&37.6&63.9&75.3&89.4\cr
% 	\multicolumn{2}{c}{DVR}      \cr
%   \multicolumn{2}{c}{DVDL}      \cr
    %\multicolumn{2}{c}{AFDA}      &37.5&62.7&73.0&81.8&43.0&72.7&84.6&91.9\cr

% 	\multicolumn{2}{c}{STFV3D+KISSME}      \cr
   
    %\multicolumn{2}{c}{STFV3D}    &37.0&64.3&77.0&86.9&42.1&71.9&84.4&91.6\cr
    \multicolumn{2}{c}{SI$^2$D\cite{zhu2016video}}  &48.7&81.1&89.2&97.3&76.7&95.6&96.7&98.9     \cr
    \multicolumn{2}{c}{TDL\cite{you2016top}}       &56.3&87.6&95.6&98.2&56.7&80.0&87.6&93.5\cr
    \midrule
    \multicolumn{2}{c}{RFA\cite{yan2016person}}  &49.3&76.8&85.3&90.0&58.2&85.8&93.4&97.9\cr
    %\multicolumn{2}{c}{RNN}       &82.77&75.90&80.74&82.84&77.56&81.89\cr
	\multicolumn{2}{c}{CNN-RNN\cite{mclaughlin2016recurrent}}      &58.0&84.0&91.0&96.0&70.0&90.0&95.0&97.0\cr
    \multicolumn{2}{c}{RCN+KISSME\cite{wu2016deep}}    
    &46.1&76.8&89.7&95.6&69.0&88.4&93.2&96.4  \cr
    \multicolumn{2}{c}{TSSCNN\cite{chung2017two}}
    &60.0&86.0&93.0&97.0&78.0&94.0&97.0&99.0\cr 
    \multicolumn{2}{c}{ASTPN\cite{xu2017jointly}}  &62.0&86.0&94.0&98.0&77.0&90.0&95.0&99.0 \cr
    \multicolumn{2}{c}{CNN+XQDA\cite{zheng2016mars}}       
    &53.0&81.4&-&95.1&77.3&93.5&95.7&99.3 \cr
	\multicolumn{2}{c}{SRM+TAM\cite{zhou2017see}}    
    &55.2&86.5&-&97.0&79.4&94.4&-&99.3  \cr
    %\multicolumn{2}{c}{CAR}
    %&60.2&85.1&-&94.2&83.3&93.3&-&96.7\cr
    \multicolumn{2}{c}{QAN*\cite{liu2017quality}}    &68.0&86.8&95.4&97.4&\bf{90.3}&\bf{98.2}&\bf{99.3}&\bf{100}\cr 
    \multicolumn{2}{c}{TRL\cite{dai2018video}}    &57.7&81.7&-&94.1&87.8&97.4&-&99.3\cr
    \midrule
    \multicolumn{2}{c}{STSRN}     &\bf{70.0}&\bf{89.3}&\bf{95.7}&\bf{98.7}&88.0&{97.0}&{99.0}&{99.0}\cr
  
    \bottomrule  
    \end{tabular*} 
    \begin{tablenotes}
	\item[-] ``*" indicates that the number of frames in tracklet is larger than 21 are used in PRID2011, which is different from common settings.
%     \item[-] The upper part of table means the hand-crafted feature, the lower part means deep learning method. Deep methods contains the state-of-the-arts from 2016 to now.
	\end{tablenotes}
    \end{threeparttable}  
\end{table}

\subsection{Comparison with the State-of-the-Art}
To further evaluate the performance of our model, we compared the proposed architecture with the state-of-the-art methods on iLIDS-VID, PRID2011 and MARS datasets.
\subsubsection{Results on iLIDS-VID and PRID2011.}
Table~\ref{tab:performance_cmc} shows results on iLIDS-VID and PRID2011 datasets. The upper part lists the state-of-the-art methods with hand-crafted features while the middle part displays recent video-based methods under the scope of deep learning. With the help of the novel spatial-temporal residual learning framework, our STSRN, listed at the bottom, outperforms all previous methods on the challenging iLIDS-VID task, and most of them on PRID2011 in terms of the CMC results. Note that our network surpasses the state-of-the-art comprehensively and transcends TRL~\cite{dai2018video} by 12.3\%, 8.4\% and 4.6\% on iLIDS-VID in terms of rank-1, rank-5 and rank-20 matching rate, which strongly demonstrates the effectiveness of our synergic residual learning across spatial domain and time domain. Compared to the QAN~\cite{liu2017quality}, our network achieves 70.0\% matching rate at rank-1, exceeding it by 2\% with much lower complexity both in network architecture and query time. Besides, our STSRN also remains substantially ahead in PRID2011 where our model set the new state-of-the-art under the common settings.
%-- except rank-1,5 we have led the other indicators. 
% And residual learning based methods have a dominant place in video person re-id tasks like QAN, TRL and STSRN at present.  

% Considering for pragmatic issue, we evaluated our model in terms of complexity of STSRN and efficiency of retrieving.
\subsubsection{Results on MARS.}
As shown in Table~\ref{tab:performance_mars}, our STSRN outperforms ASTPN~\cite{xu2017jointly} and CNN-RNN~\cite{mclaughlin2016recurrent} by a large margin on MARS. More explicitly, our STSRN surpasses ASTPN and CNN-RNN by 32.7\% and 36.7\% at CMC rank-1 matching rate, respectively, which demonstrates superior performance of our STSRN on challenging video ReID dataset again.
% , 23.8\%, 22.8\%, 17.1\%
%\vspace{-0.8cm}
\begin{table}[t]
  \centering  
  %\fontsize{6.5}{8}\selectfont  
  \begin{threeparttable}
  \caption{Comparisons on MARS in terms of CMC matching rate (\%).}    \label{tab:performance_mars}
  \begin{tabular*}{0.9\textwidth}{l@{\extracolsep{\fill}}*{6}{c}}
%     \begin{tabular}{0.75\textwidth}{cccccc}
    \toprule  
    \multicolumn{2}{c}{\multirow{2}{*}{Model}}
    &\multicolumn{4}{c}{MARS}\cr
    \cmidrule(lr){3-6} 
    &&R1&R5&R10&R20\cr  
    \midrule 
    \multicolumn{2}{c}{CNN-RNN\cite{mclaughlin2016recurrent}}  &40.0&60.0&70.0&77.0\cr
    \multicolumn{2}{c}{ASTPN\cite{xu2017jointly}}  &44.0&70.0&74.0&81.0\cr
    \multicolumn{2}{c}{STSRN}  &\bf{76.7}&\bf{93.8}&\bf{96.8}&\bf{98.1}\cr
    \bottomrule
    \end{tabular*} 
%     \begin{tablenotes}
% 	\end{tablenotes}
    \end{threeparttable}
    %\centering
\end{table}
%

%\vspace{-0.8cm}
\subsubsection{Complexity}
As far as computing saving is concerned, our model is superior to competitor methods about the number of parameters;for a better illustration of this, Table~\ref{table:parameter} directly quantifies the previous state-of-the-arts and ours. It could be easily concluded that STSRN has a comprehensive advantage over others, and its parameters are only one tenth of GoogLeNet~\cite{szegedy2015going}.
Compared to previous research that usually strike a balance between the model complexity and characterization capability, our model is both light-weighted and high-performing network, mainly by benefiting from the synergic residual learning.
% Within 590k parameters, we have a more satisfactory performance{略高一些的水平}. Obviously we did better at complexity.
%Although all of them obtain similar capability of feature presentation, both the QAN and the TRL adopted an very deep networks, GoogleNet, and QAN had a scorer while TRL processed the temporal representation by Bi-LSTM, which obviously results in much more parameters than our STSRL network only containing double small-scale residual blocks and one-directional RNN, without any extra branches.
\begin{table}[t]
\centering
\caption{Comparisons on the number of parameters of models}
\label{table:parameter}
\begin{tabular}{|c|c|c|c|c|c|}
\hline
Models     & ~CNN-RNN\cite{mclaughlin2016recurrent}~   & ~TSSCNN\cite{chung2017two}~  & ~QAN\cite{liu2017quality}~  & ~TRL\cite{dai2018video}~& ~Ours~    \\ \hline
~Parameters~ &430k & 860k & 6.7M &$\approx$30M & 590k \\ \hline
\end{tabular}
\end{table}

\subsubsection{Efficiency.}
Most model with attention mechanism~\cite{zhou2017see,xu2017jointly} suffer from grossly inefficient computation because 
of the mutual influence between the query and the gallery. 
% In other words, 
% they need to
% consider the attention score for every pair.
%correlation probabilities $\{p^{(i)}_1,...,p^{(i)}_k\}$ between probe $q_i$ and gallery $\{g_1,g_2,...,g_k\}$.  
%In our model, we obviate this inefficient dependent computing. owing to we take no account of the dependent computing
However, our systerm employs pre-computing tactic, i.e., features of identities are extracted as the memory. When a query comes, the systerm merely rank the distance between the memory and current query. Suppose we have $X$ examples to query and $Y$ examples in the gallery. The attention-based approaches need to extract features for $XY$ times while we only need $X + Y$ times.
% Every retrieving, .
%For instance, suppose we have $m$ examples to query and $k$ examples in the gallery. During every example retrieving, we apply $k$ times distance metric operation based on features. We just need to extract $m+k$ features and only one feature per retrieving to complete the retrieval task.  However, it is impossible for the attentive models to ignore the correlation of $m$ and $k$. Thus they have to extract $m \times k$ features with $k$ features per retrieving. 

\subsection{Cross-Dataset Generalization}
Due to the variety of geometric and environmental conditions, models trained on one dataset maybe exhibit poor performance on another, due to the over-fitting trap. 
To better evaluate the generality of our model, we conducted cross-dataset experiments which were trained on iLIDS-VID and tested on PRID2011, showing the results in Table~\ref{tab:CrossDataset}. We achieve 32.0\%, 58.0\%, 71.0\% and 90.0\% of the CMC scores at rank-1, 5, 10 and 20,  exceeding all baselines except for being slightly inferior to TRL method at rank-5, which proves certain generality of our model.

\begin{table}[t]
\caption{Cross-Dataset Testing on PRID2011 in terms of CMC rank1/5/10/20}
\label{tab:CrossDataset}
\begin{center}
\begin{tabu} to 0.8\textwidth{X[3,c]|X[2,c]|X[2,c]|X[2,c]|X[2,c]}  
%0.8\textwidth   为设置表格宽度
%X[c]      表示这一列居中，所占比例为1，相当于X[1,c]  
%X[3,c]   表示这一列居中，所占比例为3，这列的宽度是X[c]列的3倍  
\hline
Method  &R1  &R5 &R10 &R20 \\  
\hline
CNN-RNN~\cite{mclaughlin2016recurrent}  &28.0   &57.0  &69.0      &81.0\\  
ASTPN~\cite{xu2017jointly}    &30.0   &58.0 &71.0      &85.0\\
TRL~\cite{dai2018video}    &29.5  &59.4   &-      &82.2\\
STSRL     &32.0   &58.0    &71.0  &90.0\\ 
% 5    &1.06$\sim$1.20  &37  &12   &2.04$\sim$2.18      &3\\  
% 6    &1.20$\sim$1.34  &53  &13   &2.18$\sim$2.38      &1\\  
% 7    &1.34$\sim$1.48  &56  &     &                    & \\  
\hline
\end{tabu}
\end{center}  
\end{table}  

\subsection{Effectiveness of Spatial Residual Blocks}
\begin{figure}[t]
\centering
\includegraphics[width=12cm]{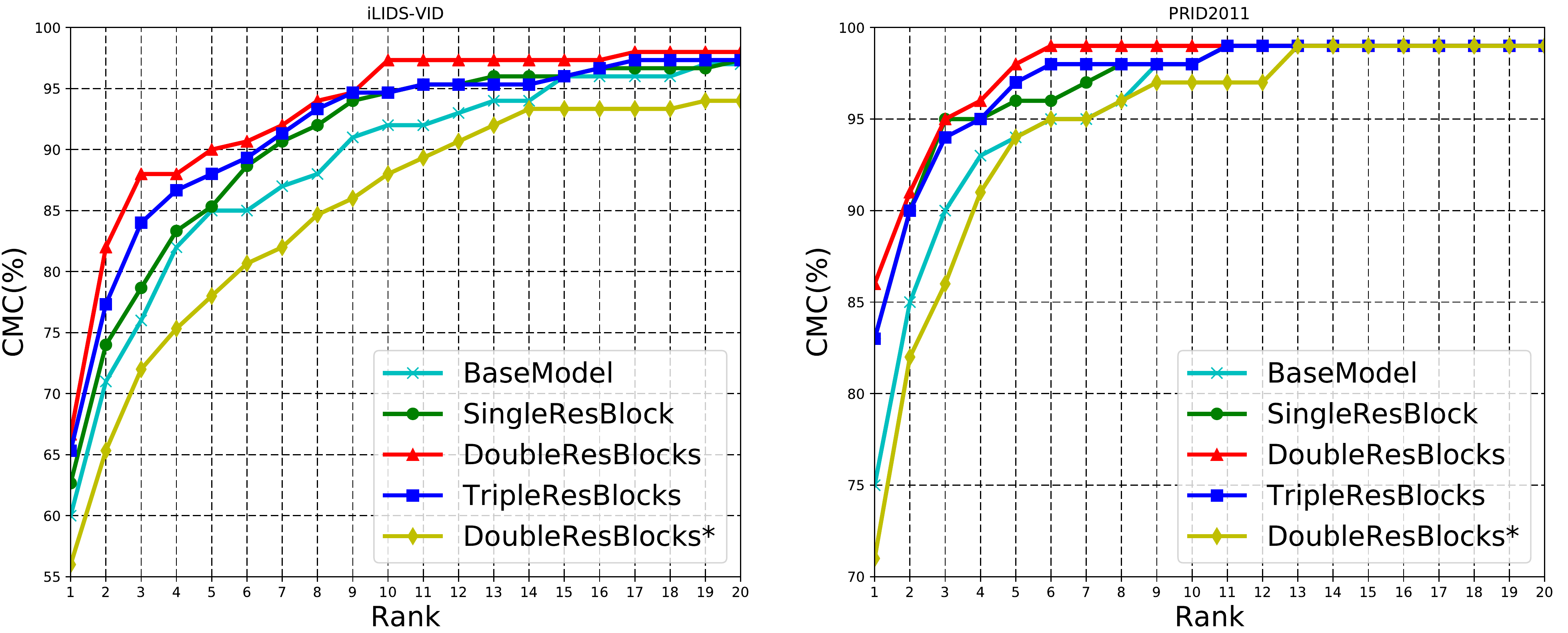}
\caption{CMC curves of different numbers and different kinds of spatial residual blocks on (a) iLIDS-VID and (b) PRID2011 datasets}
\label{fig_spatialresult}
\end{figure}

As shown in Figure~\ref{fig_3dmodel}, our spatial submodule contains double residual blocks. Inspired by the strong evidence shown in~\cite{he2016deep}, we also expected that the residual learning principle could be generalized to our re-ID task. 
We performed experiments mainly related to the number and form of spatial residual blocks. The description of experimental models and their CMC curves from rank-1 to rank-20 are illustrated in Table~\ref{tab:combinedSpatialBlocks} and Figure~\ref{fig_spatialresult}, respectively.

From the CMC curves in Figure~\ref{fig_spatialresult}, 
we have three important observations. First, the introduction of adequate residual blocks in the form of Figure~\ref{fig_resblock}c can significantly boost the CMC accuracy from rank-1 to rank-20. Specifically, both {\em DoubleResBlocks} and {\em SingleResBlock} surpass {\em BaseModel} without any residual connections by a large margin for iLIDS-VID. Besides, the rank-1 matching rate of {\em DoubleResBlocks} and {\em SingleResBlock} for PRID2011 outperforms that of {\em BaseModel} by 11.0\% and 8.0\%, respectively, and even for the more challenging iLIDS-VID, the rank-1 accuracy is improved by about 6.7\% when {\em BaseModel} changes to {\em DoubleResBlocks}. Second, it is not better to simply add more residual blocks. The CMC curve of {\em TripleResBlocks} performs worse than {\em DoubleResBlocks} on two datasets in highest ranks. Third, the rank-1 accuracies fall down to just 56.3\% and 71.0\% for iLIDS-VID and PRID2011, respectively when the architecture moves from Figure~\ref{fig_resblock}c to Figure~\ref{fig_resblock}a, which demonstrates that larger receptive filed and the max pooling are more suitable for our tasks.

In addition, when we replaced the residual blocks of {\em DoubleResBlocks} with Figure~\ref{fig_resblock}b, the CMC scores at rank-1 was stuck at 63.3\% for iLIDS-VID, which is inferior to that of {\em TripleResBlocks}. It reflects the gradient vanishing problem caused by \emph{Tanh}.
% the residual blocks perform a linear projection by the shortcut connections with greater discriminative power

% In addition, ???ResNet but only got 62, 88, 92, 98 on 
% Based on these facts, we can deduce that double res blocks are more applicable to video person re-id than other models.
% reasonable network topologies
% The degradation of training accuracy indicates that not all systerms are similarly easy to optimize. With the network going deeper, accuracy gets saturated and then degrades rapidly. 
%Our current solvers on hand are unable to find solutions that are comparably good or better than the constructed solution(or unable to do so in feasible time).

\begin{table}[t]
\caption{Comparable models for exploring how to make full use of spatial residual connections. All models are followed by a fully-connected layer and the vanilla recurrent neural network. The ``Conv'', ``Res'' and ``Res*'' are the convolutional layer, the residual block of Figure~\ref{fig_resblock}c and Figure~\ref{fig_resblock}a, respectively, which are all followed by a Tanh layer. The kernel size, padding and stride of ``Conv"s are set to 5, 4 and 1, respectively. The ``Max" is the max pooling layer with window size 2 and stride 2}
\label{tab:combinedSpatialBlocks}
\begin{center}
\begin{tabu} to 1\textwidth{|X[3,c]|X[c]|X[c]|X[c]|X[c]|X[c]|X[c]|X[c]|X[c]|}  
%0.8\textwidth   为设置表格宽度
%X[c]      表示这一列居中，所占比例为1，相当于X[1,c]  
%X[3,c]   表示这一列居中，所占比例为3，这列的宽度是X[c]列的3倍  
\hline
% \multicolumn{n}{fixup}{text}
% Dataset  &\multicolumn{4}{c|}{iLIDS-VID}&\multicolumn{4}{c}{PRID2011}\\
% \hline
Model  &Layer&Max&Layer&Max &Layer&Max&Layer&Max\\
\hline
BaseModel  &Conv &+ &Conv &+ &Conv &+ &- &-\\
\hline
SingleResBlock  &Conv &+ &Conv &+ &Res &+ &- &-\\
\hline
DoubleResBlocks  &Conv &+ &Res &+ &Res &+ &- &-\\
\hline
TripleResBlocks  &Conv &+ &Res &- &Res &+ &Res&+\\
\hline
DoubleResBlocks*  &Conv &+ &Res* &+ &Res* &+ &- &-\\
\hline
\end{tabu}
\end{center}  
\end{table}
% \begin{enumerate}
% \item BaseModel: 3 $\times$ Conv
% \item SingleResBlock: 2 $\times$ Conv $\to$ Res Block
% \item DoubleResBlocks:  Conv  $\to$ 2 $\times$ Res Block
% \item TripleResBlocks:  Conv  $\to$ Res Block $\to$ 2 $\times$ Res Block
% \item DoubleResBlocksS2:  Conv  $\to$ 2 $\times$ Res Block
% \end{enumerate}

% \begin{figure}[t]
% \centering
% \includegraphics[width=12cm]{SpatialResBlocks.pdf}
% \caption{CMC curves of different numbers and different kinds of spatial residual blocks on (a) iLIDS-VID and (b) PRID2011 datasets}
% \label{fig_spatialresult}
% \end{figure}

\begin{table}[t]
\caption{CMC scores in terms of rank-1, rank-5, rank-20 for {\em BaseModel} and {\em DoubleResBlocks} with/without residual RNN or STSM on iLIDS-VID and PRID2011 datasets. The ``TemRes'' and ``STSM'' mean temporal residual learning and the Spatial-Temporal Smooth Module, respectively}
\label{tab:TemResAndSTSM}
\begin{center}
\begin{tabu} to 1\textwidth{X[3,c]|X[c]|X[c]|X[c]|X[c]|X[c]|X[c]}  
%0.8\textwidth   为设置表格宽度
%X[c]      表示这一列居中，所占比例为1，相当于X[1,c]  
%X[3,c]   表示这一列居中，所占比例为3，这列的宽度是X[c]列的3倍 
\hline
% \multicolumn{n}{fixup}{text}
Dataset  &\multicolumn{3}{c|}{iLIDS-VID}&\multicolumn{3}{c}{PRID2011}\\
\hline
Rank  &1&5&20 &1&5&20\\
\hline
BaseModel  &60.3 &87.7 &96.7 &75.0 &95.0 &98.0\\
\hline
DoubleRes  &66.7 &89.3 &97.3 &86.0 &96.0 &99.0\\
\hline
Base+TemRes  &61.3 &88.3 &97.7 &78.0 &93.0 &99.0\\
\hline
Double+TemRes  &68.7 &90.0 &98.3 &87.0 &97.0 &99.0\\
\hline
Base+TemRes+STSM &63.7 &87.3 &98.7 &80.0 &96.0 &99.0\\
\hline
Double+TemRes+STSM  &70.0 &89.3 &98.7 &88.0 &97.0 &99.0\\
% $Method$  &\multicolumn{2}{c}{text}&\multicolumn{2}{c}{text}\\
% &$R10$ &$R20$ \\  
% ASTPN~\cite{xu2017jointly}    &30.0   &58.0 &71.0      &85.0\\
% TRL~\cite{dai2018video}    &29.5  &59.4   &-      &82.2\\
% Ours     &32.0   &58.0    &71.0  &90.0\\ 
% 5    &1.06  &37  &12   &2.04 &&&&    \\  
% 6    &1.20$\sim$1.34  &53  &13   &2.18$\sim$2.38      &1\\  
% 7    &1.34$\sim$1.48  &56  &     &                    & \\  
\hline
\end{tabu}
\end{center}  
\end{table}
\subsection{Evaluation on Temporal Residual Learning}
\begin{figure}[t]
\centering
\includegraphics[width=9cm,height=5cm]{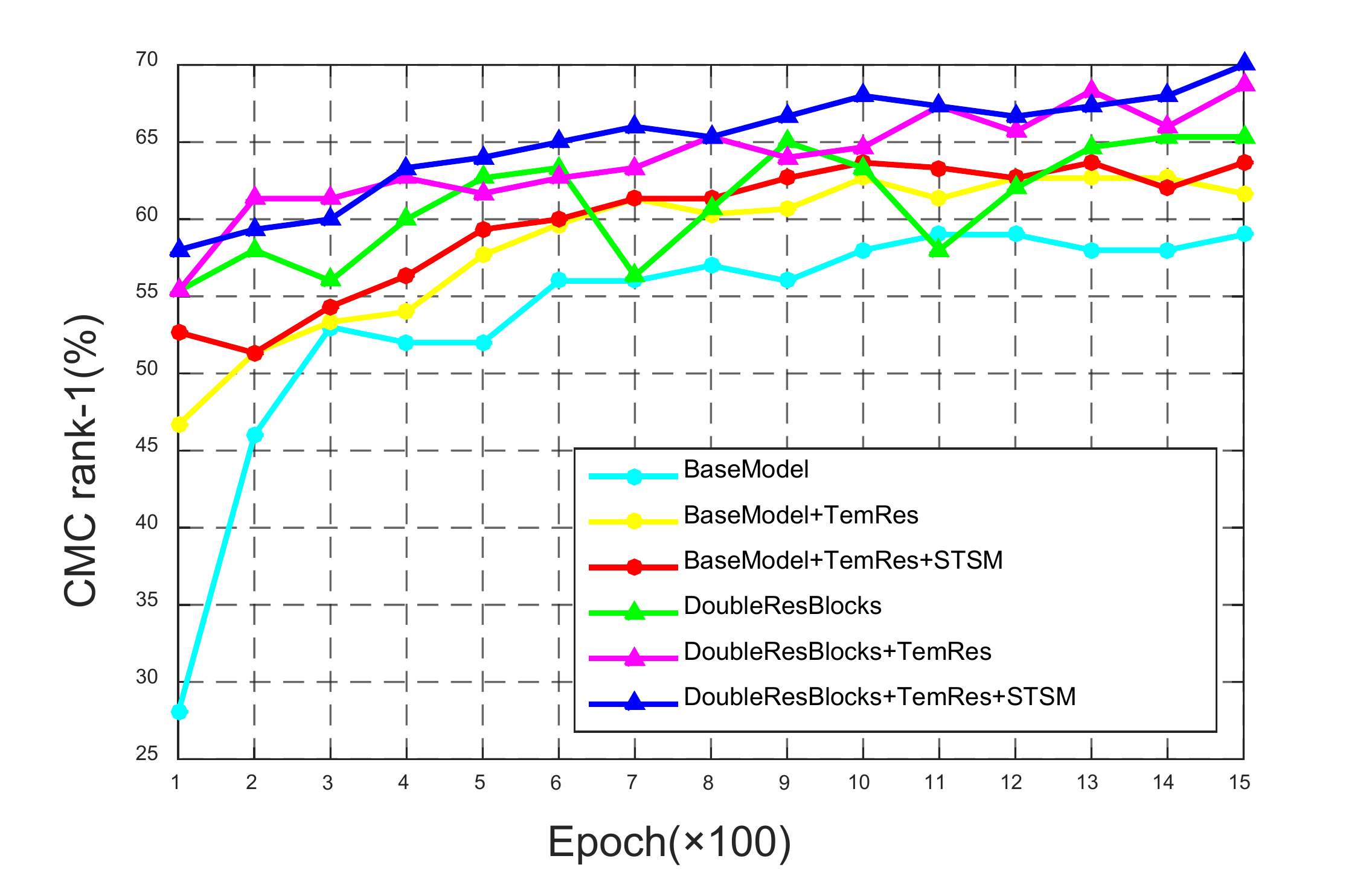}
\caption{CMC scores at rank-1 with the progress of training for {\em BaseModel} and {\em DoubleResBlocks} models with/without residual RNN or STSM on iLIDS-VID. The ``TemRes'' and ``STSM'' means temporal residual learning and the Spatial-Temporal Smooth Module, respectively}
\label{fig_TemResAndSTSM}
\end{figure}

Due to significant viewpoint/illumination variations as well as background clutter and occlusions, a good video-based re-ID method should grab these diversities selectively, especially the temporal cues. To illustrate how our temporal residual learning works, we display the rank-1 matching rate curves with the progress of training in Figure~\ref{fig_TemResAndSTSM}, and record CMC rank-1, rank-5, rank-20 scores at Table~\ref{tab:TemResAndSTSM}.

From Figure~\ref{fig_TemResAndSTSM}, it can be easily observed that the residual learning in RNN brings considerable performance gains along with the powerful ability of noise resistance for both {\em BaseModel} and {\em DoubleResBlocks} during the training phase. Specifically, {\em BaseModel} only reaches an accuracy of 28.3\% in the $100^{th}$ training epoch while {\em BaseModel+TemRes} greatly improves on that, achieving an accuracy of 46.7\%. More clearly, the {\em DoubleResBlocks+TemRes} substantially shows smaller jitter and possesses slightly higher accuracy than its counterpart overall.

From Table~\ref{tab:TemResAndSTSM}, we can see that the temporal residual learning improves CMC scores for {\em BaseModel} on both datasets. With the help of the residual RNN, there are 1.0\% and 3.0\% lift at rank-1 score on iLIDS-VIDS and PRID2011, respectively for {\em BaseModel} while 2.0\% and 1.0\% for {\em DoubleResBlocks}.

In a word, our temporal residual processor actually helps to prevent redundant information from worsening discriminative feature extraction. In other words, the models equipped with residual connections temporally grab diversities across video sequence selectively, making it more robust to noisy features.

\subsection{Discussion about STSM}
As illustrated in Section 3.3, we devise the STSM to smooth noisy motion in adjoining frames. Here we compare models whether assembled with STSM or not at Table~\ref{tab:TemResAndSTSM}. More detailed tendency of the CMC curves are shown in Figure~\ref{fig_TemResAndSTSM}. As we expect, models with STSM outperform ones without STSM. For example, {\em BaseModel+TemRes+STSM} is about 2\% higher than {\em BaseModel+TemRes} on both datasets while there are 1.3\%, 1.0\% improvements for {\em DoubleRes} in terms of rank-1 matching rate on iLIDS-VID, PRID2011, respectively. These evidence experimentally illustrate that STSM can effectively enhance spatial-temporal synergic residual learning.

\section{Conclusions}
In this paper we develop a succinct but powerful framework, Spatial-Temporal Synergic Residual Network (STSRN), for video-based person re-identification. Our spatial residual blocks effectively extract spatial features for single image and the recurrent residual learning module provides alternative signals. Besides, the parameter-efficient spatial-temporal smooth module (STSM) can further improve the robustness of the model. Both extensive experiments results and detailed analyses on MARS, iLIDS-VID and PRID2011 datasets strongly demonstrate that STSRN breaks through the bottlenecks in small spatial-temporal networks for person re-id. STSRN would be adapted to person detection and tracking tasks in the future \cite{conf/cvpr/WangCF16}.
 % help us to explore the most reasonable topology.

%\clearpage

\bibliographystyle{splncs}
\bibliography{egbib}
\end{document}